\documentclass{article} % For LaTeX2e
\usepackage{iclr2020_conference,times}

\usepackage{amsmath,amsfonts,bm}

\def\eqref#1{equation~\ref{#1}}
\def\ceil#1{\lceil #1 \rceil}

\def\1{\bm{1}}

\DeclareMathAlphabet{\mathsfit}{\encodingdefault}{\sfdefault}{m}{sl}
\SetMathAlphabet{\mathsfit}{bold}{\encodingdefault}{\sfdefault}{bx}{n}

\usepackage{hyperref}
\usepackage{url}
\usepackage{multicol}
\usepackage{booktabs}
\usepackage{multirow}
\usepackage{subcaption}
\usepackage{mathtools}
\usepackage{tablefootnote}
\usepackage{color}
\iclrfinaltrue
\usepackage{pgfplotstable} %tikz plots data
\usepackage{pgfplots}
\pgfplotsset{compat=1.14}
\makeatletter
\newcommand{\ssymbol}[1]{$^{\@fnsymbol{#1}}$}
\makeatother

\title{MUSE: Parallel Multi-Scale Attention for Sequence to Sequence Learning}

\author{Guangxiang Zhao\ssymbol{2}, Xu Sun\ssymbol{2}\ssymbol{3}\thanks{ Corresponding author} , Jingjing Xu\ssymbol{3}, Zhiyuan Zhang\ssymbol{3}, Liangchen Luo\ssymbol{3} \\
\ssymbol{2}Center for Data Science, Peking University\\
\ssymbol{3}MOE Key Lab of Computational Linguistics, School of EECS, Peking University\\
\texttt{\{zhaoguangxiang,xusun,jingjingxu,zzy1210,luolc\}@pku.edu} \\
}

\begin{document}

%Experimental results also show that the performance drops largely in the late training process.
\maketitle

% parallels the self-attention module with convolutional modules  in each layer. Convolutional modules with different kernel sizes can capture multi-grained features, thus improving the robustness of Transformer.

%The success is inseparable with the attention mechanism that can capture long-distance dependencies. In this paper, we rethink a fundamental question: \textsl{Is attention  indeed all you need?} %Transformers, solely based an attention mechanism, have achieved state-of-the-art results on a variety of natural language processing tasks. Transformer, a method that is solely based an attention mechanism, have achieved state-of-the-art results on a variety of natural language processing tasks.
\begin{abstract}
% In this work, we rethink a fundamental question: ``Is attention really all you need?''.  
% Although self-attention has achieved superior results on sequence to sequence learning tasks over convolutional models, the  self-attention mechanism alone is still weak in long sequence modeling. 
% % The global attention map is too dispersed to capture valuable information. In such case, the local/token features that are also significant to sequence modeling are omitted to some extent. 
% In this work we refuse to make choices in self-attention and convolution, and explore parallel multi-scale representation learning on sequence data. We propose a new module, Parallel \textbf{ MU}lti-\textbf{S}cale att\textbf{E}ntion (\textbf{MUSE}).  The new module combines convolution and self-attention in  one  module for sequence to sequence learning. It first encodes the input into hidden representations, and then performs  self-attention and convolution transformations at the same hidden space in parallel.
% Experimental results show that 
% the proposed model achieves substantial performance improvements over Transformer, especially on long sequences, and outperforms all previous models with the comparable model size and the same training data  on three main machine translation tasks. To be specific, we achieve a BLEU score of 36.3 on IWSLT De-En,  29.9 on WMT14 En-De, and 43.5 on WMT14 En-Fr. 
% In addition, it also opens a new direction on parallel sequence representation learning, which leads to efficient training and inference. 
% MUSE has potential for accelerating training and inference.

In sequence to sequence learning, the self-attention mechanism proves to be highly effective, and achieves significant improvements in many tasks. However, the self-attention mechanism is not without its own flaws. Although self-attention can model extremely long dependencies, the attention in deep layers tends to over-concentrate on a single token, leading to insufficient use of local information and difficultly in representing long sequences. In this work, we explore parallel multi-scale representation learning on sequence data, striving to capture both long-range and short-range language structures. To this end, we propose the Parallel \textbf{ MU}lti-\textbf{S}cale att\textbf{E}ntion (\textbf{MUSE}) and MUSE-simple. MUSE-simple contains the basic idea of parallel multi-scale sequence representation learning, and it encodes the sequence in parallel, in terms of different scales with the help from self-attention,  and pointwise transformation. MUSE builds on MUSE-simple and explores  combining convolution and self-attention for learning sequence representations from more different scales.
We focus on machine translation and the proposed approach achieves substantial performance improvements over Transformer, especially on long sequences. More importantly, we find that although conceptually simple, its success in practice requires intricate considerations, and the multi-scale attention must build on unified semantic space. Under common setting, the proposed model achieves substantial performance and outperforms all previous models on three main machine translation tasks. In addition, MUSE has potential for accelerating inference due to its parallelism.
Code will be available at \url{https://github.com/lancopku/MUSE}.

\end{abstract}
\section{Introduction}
% In recent years, Transformer has been a widely-used text sequence modeling network due to its promising results on a variety of natural language processing tasks,
In recent years, Transformer has been remarkably adept at sequence learning tasks like machine translation~\citep{vaswani2017attention,dehghani2018universal}, 
% summrization~\citep{liu2019text},
text classification~\citep{devlin2018bert,yang2019xlnet}, language modeling~\citep{sukhbaatar2019augmenting,Dai_2019}, etc. It is solely based on an attention mechanism that captures global dependencies between input tokens, dispensing with recurrence and convolutions entirely. The key idea of the self-attention mechanism is updating token representations based on a weighted sum of all input representations.  

%The self-attention mechanism, which directly captures global structures, shows significant performance gains over convolutional models, which can only capture local structures at each time. 
However, recent research~\citep{DBLP:conf/emnlp/TangMRS18} has shown that the Transformer has surprising shortcomings in long sequence learning, exactly because of its use of self-attention. As shown in Figure 1 (a), in the task of machine translation, the performance of Transformer drops with the increase of the source sentence length, especially for long sequences. The reason is that the attention can be over-concentrated and disperse, as shown in Figure 1 (b), and only a small number of tokens are represented by attention. It may work fine for shorter sequences, but for longer sequences, it  causes insufficient representation of information and brings difficulty for the model to comprehend the source information intactly. In recent work, local attention that constrains the attention to focus on only part of the sequences \citep{child2019generating,Sukhbaatar_2019} is used to address this problem. However, it costs self-attention the ability to capture long-range dependencies and also does not demonstrate effectiveness in sequence to sequence learning tasks.

% Despite  significant performance of self-attention over convolutional models, recent research has accused the Transformer model of shortcomings in long-sequence modeling~\citep{DBLP:conf/emnlp/TangMRS18}. As shown in Figure 1 (a),  the performance of Transformer drops largely with the increase of the source sentence length. Based on the empirical analysis, we find that the performance drop is mainly due to the dispersed attention map where local/token features, also significant to sequence modeling, are omitted to some extent, as shown in Figure 1 (b).  Although local attention enables long sequence modeling in recent work \citep{child2019generating,Sukhbaatar_2019}, they not only do not demonstrate effectiveness in sequence to sequence learning tasks, but also add local context constraints to self-attention that may not be true if there exists dependency at long time scale.
% Although the original Transformer encodes position information into token embeddings to address this problem, it is still unknown how much distance information is kept in the token representation with the increase of layer depth.

% child2019generating
\begin{figure}[ht]
\centering
\includegraphics[width=0.46\textwidth]{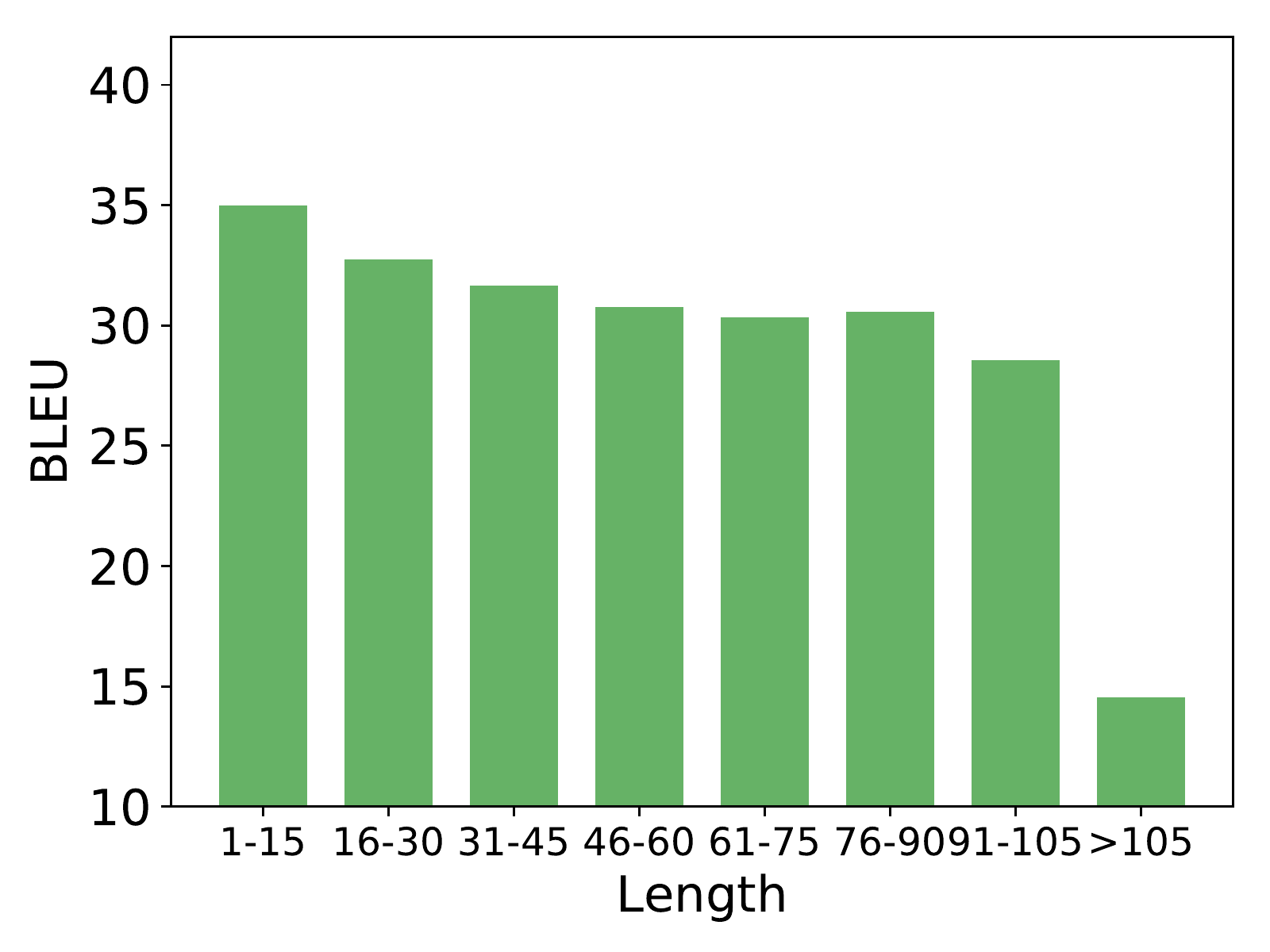} \qquad
\includegraphics[width=0.48\textwidth]{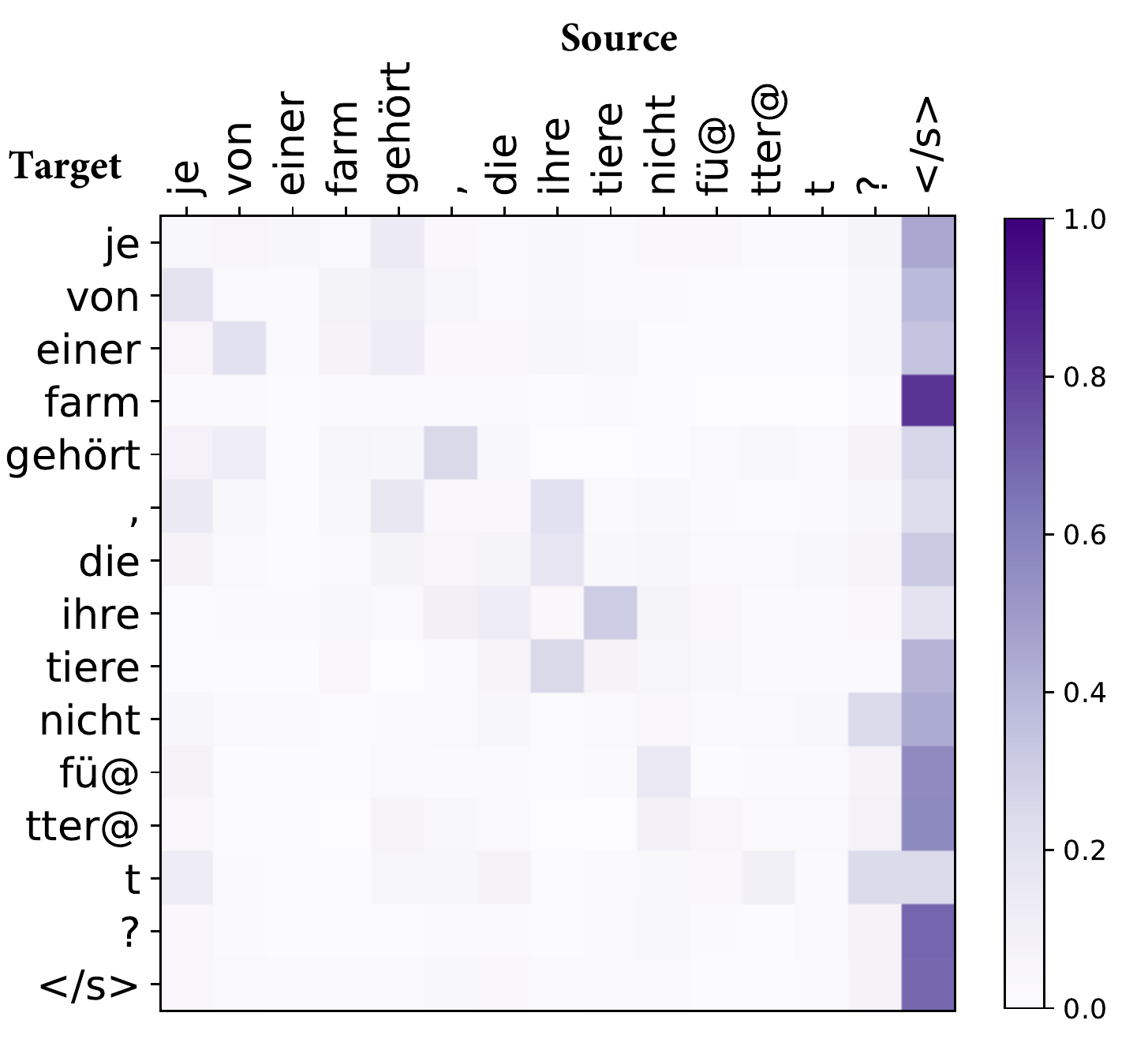}
\caption{The left figure shows that the performance drops largely with the increase of sentence length on the De-En dataset. The right figure shows the attention map from the $3$-th encoder layer. As we can see, the attention map is too dispersed to capture sufficient information. For example, ``[EOS]'', contributing little to word alignment, is surprisingly over attended.}
\label{fig:motivation}
\end{figure}

To build a module with both inductive bias of local and global context modelling in sequence to sequence learning, we hybrid self-attention with convolution and present Parallel multi-scale attention called MUSE.  It  encodes inputs into hidden representations and then  applies self-attention  and  depth-separable convolution transformations in parallel. The convolution compensates for the insufficient use of local information while the self-attention focuses on capturing the dependencies. Moreover, this parallel structure is highly extensible, and new transformations can be easily introduced as new parallel branches, and is also favourable to parallel computation.

% This module has a very strong scalability, token level transformation and context attention can also be computed in parallel.

The main contributions are summarized as follows:
\begin{itemize}

\item We find that the attention mechanism alone suffers from dispersed weights and is not suitable for long sequence representation learning. The proposed method tries to address this problem and achieves much better performance on generating long sequence.

\item We propose a parallel multi-scale attention and explore a simple but efficient method to successfully combine convolution with self-attention all in one module. 
% Experiments establish that multi-scale attention is important for sequence to sequence learning.

\item MUSE outperforms all previous models with same training data and the comparable model size, with state-of-the-art BLEU scores on three main machine translation tasks. 

\item MUSE-simple introduce parallel representation learning  and brings expansibility and parallelism. Experiments show that the inference speed can be increased by 31\% on GPUs.
\end{itemize}

\begin{figure}[t]
\centering
\includegraphics[width=0.8\textwidth]{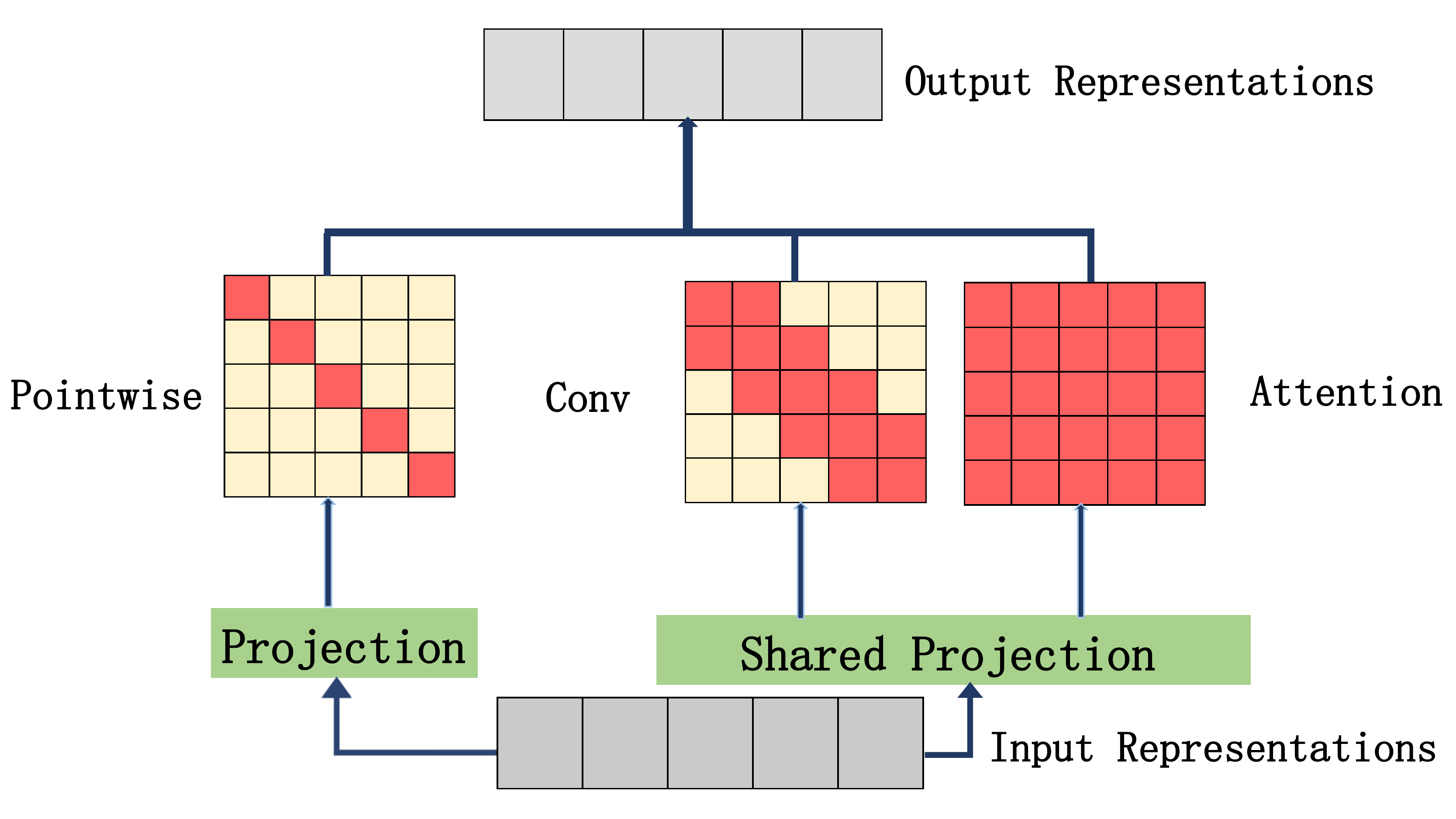}
\caption{Multi-scale attention hybrids point-wise  transformation, convolution, and self-attention to learn multi-scale sequence representations in parallel. We project convolution and self-attention into the same space to learn contextual representations.}
\label{fig:ffn_conv}
\end{figure}

\section{MUSE: Parallel Multi-Scale Attention}
%We propose a new multi-scale attention model to fuse token features, local content, and global dependencies. 

%\subsection{Overview}

%Our model is built on a Transformer network. We first review the details of Transformer in this section. 

Like other sequence-to-sequence models, MUSE also adopts an encoder-decoder framework. The encoder takes a sequence of word embeddings $(x_1, \cdots, x_n)$ as input where $n$ is the length of input. It transfers word embeddings to a sequence of hidden representation $\boldsymbol{z} = (z_1, \cdots, z_n)$. Given $\boldsymbol{z}$, the decoder is responsible for generating a sequence of text $(y_1, \cdots, y_m)$ token by token. 

The encoder is a stack of $N$  MUSE modules. Residual mechanism and layer normalization are used to connect two adjacent layers.  
The decoder is similar to encoder, except that each MUSE module in the decoder not only captures features from the generated text representations  but also performs attention over the output of the encoder stack through additional context attention. Residual mechanism and layer normalization are also used to connect two modules and two adjacent layers. 

% The decoder is also a stack of $N$ layers. Each layer contains two sub-layers: a MUSE module and a context attention module. The MUSE module is responsible for capturing features from the generated text representations. The context-attention performs attention over the output of the encoder stack.   Residual mechanism and layer normalization are also used to connect two modules and two adjacent layers. 
%The context-attention projects the encoder stack into key $K$, value $V$, and projects the output of MUSE sub-layer into query $Q$. Then, it performs the same operation as self-attention to generate the output representations. The context-attention projects the encoder stack into key $K$, value $V$, and projects the generated text representation into query $Q$. Then, it performs the same operation as self-attention to generate the output representations.

%\subsection{Multi-Scale Feature Fusion}
The key part in the proposed model is the MUSE module, which contains three main parts: self-attention for capturing global features, depth-wise separable convolution for capturing local features, and a position-wise feed-forward network for capturing token features. The module takes the output of $(i-1)$ layer as input and generates the output representation in a fusion way:
\begin{equation}
%    X_{i} = MultiHead(X_{i-1}) + Conv(X_{i-1}) + FFN(X_{i-1})
    X_{i} = X_{i-1} + Attention(X_{i-1}) + Conv(X_{i-1}) + Pointwise(X_{i-1})
%\label{layer}
\end{equation}
where  ``Attention'' refers to self-attention, ``Conv'' refers to dynamic convolution, ``Pointwise'' refers to a position-wise feed-forward network. The followings list the details of each part. %We starts from self-attention mechanism.

We also propose MUSE-simple, a simple version of MUSE, which generates the output representation similar to the MUSE model except for that it dose not the include convolution operation:
\begin{equation}
%    X_{i} = MultiHead(X_{i-1}) + Conv(X_{i-1}) + FFN(X_{i-1})
    X_{i} = X_{i-1} + Attention(X_{i-1}) + Pointwise(X_{i-1})
%\label{layer}
\end{equation}

\subsection{Attention Mechanism for Global Context Representation}

%\paragraph{Self-attention} 
Self-attention is responsible for  learning representations of global context.
 For a given input sequence $X$, it first projects $X$ into three representations, key $K$, query $Q$, and value $V$. Then, it uses a self-attention mechanism to get the output representation:
\begin{equation}
\begin{split}
 %MultiHead(X) & = Concat(head_1, \cdots, head_m)W^O   \\
 Attention(X) & =  \sigma(QW^Q,KW^K, VW^V)W^O   \\
\text{where}\ \ \ Q, K, V & = Linear_1(X), \ Linear_2(X), \ Linear_3(X) 
\end{split}
\end{equation}
Where $W^O$, $W^Q$, $W^K$, and $W^V$ are projection parameters. The self-attention operation $\sigma$ is the dot-production between key, query, and value pairs:
\begin{equation}
 \sigma(Q_1, K_1, V_1) = \text{softmax}(\frac{Q_1K_1^T}{\sqrt{d_k}})V_1
\end{equation}

Note that we conduct a projecting operation over the value in our self-attention mechanism $V_1=VW^V$ here.

\subsection{Convolution for Local Context Modeling} 
%\textbf{Convolution Augmented  MUSE}
We introduce convolution operations into MUSE to capture local context. To learn contextual sequence representations in the same hidden space, we choose depth-wise convolution~\citep{Chollet_2017} (we denote it as DepthConv in the experiments) as the convolution operation because it includes two separate transformations, namely, point-wise projecting transformation and contextual transformation. It is because that original convolution operator is not separable, but DepthConv can share the same point-wise projecting transformation with  self-attention mechanism. We choose dynamic convolution~\citep{wu2019pay}, the best variant of DepthConv, as our implementation. 

Each convolution sub-module contains multiple cells with different kernel sizes. They are used for capturing different-range features. The output of the convolution cell with kernel size $k$ is:
\begin{equation}
\begin{split}
    Conv_{k}(X) &= Depth\_conv_{k}(V_2)W^{out} \\
     V_2 &= XW^{V}
\end{split}
\end{equation}
where $W^{V}$ and $W^{out}$ are parameters, $W^{V}$ is a point-wise projecting transformation matrix. The $Depth\_conv$ refers to depth convolution in the work of  \citet{wu2019pay}. For an input sequence $X$, the output $O$ is computed as: 
\begin{equation}
    O_{i,c} = Depth\_conv_{k}(X) = \sum\limits_{j=1}^{k} \big( \text{softmax} (\sum\limits_{c=1}^d W_{j,c}^Q X_{i,c})  \cdot X_{i+j-\ceil{\frac{k+1}{2}}, c} \big)
\end{equation}
where $d$ is the hidden size.
%we refer this model as MUSE-FF-CNN(w/o wt).
Note that we conduct the same projecting operation over the input in our convolution mechanism $V_2=XW^V$ here with that in self-attention mechanism. 

\textbf{Shared projection}
To learn contextual sequence representations in the same hidden space, the projection in the self-attention mechanism $V_1=VW_V$ and that in the convolution mechanism $V_2=XW^V$ is shared. Because the shared projection can project the input feature into the same hidden space. If we conduct two independent projection here: $V_1=VW_1^V$ and $V_2=XW^V_2$, where $W_1^V$ and $W_2^V$ are two parameter matrices, we call it as separate projection. We will analyze the necessity of applying shared projection here instead of separate projection.

\textbf{Dynamically Selected Convolution Kernels}
 We introduce a gating mechanism to automatically select the weight of different convolution cells. 
\begin{equation}
    Conv(X) = \sum\limits_{i=1}^{n} \frac{\exp{(\alpha_i)} }{\sum\limits_{j=1}^{n}\exp{(\alpha_j)}} {Conv_{k_{i}}(X)}
\end{equation}
% where $\alpha_i$ is a scalar initialized with $1/n$. $n$ is the number of cells.

% Then we add the add the representations of the  dynamic convolution to the main representations before the residual connection.
% \begin{equation}
%     X_{i} = LayerNorm(MultiHead(X_{i-1}) + Conv(X_{i-1}) + FFN(X_{i-1}) + X_{i-1})
% \label{layer}
% \end{equation}

\subsection{Point-wise Feed-forward Network for Capturing Token Representations} %   add mha and ffn % ffn is k=1, and mha is k=n  
To learn  token level representations, MUSE concatenates an self-attention network  with a position-wise feed-forward network at each layer.  Since the linear transformations are the same across different positions, the position-wise feed-forward network can be seen as a token feature extractor. % Considering parameter size, we put the position-wise feed-forward network in the original Transformer at the same layer with the attention to re-use its parameter to capture token features in MUSE:
\begin{equation} 
%FFN(x) = max(0, H_lW_1 + b_1)W_2 + b_2
Pointwise(X) = max(0, XW_1 + b_1)W_2 + b_2
\end{equation}
where $W_1$, $b_1$, $W_2$, and $b_2$ are projection parameters.
\section{Experiment}
We evaluate MUSE on four machine translation tasks. This section describes the datasets, experimental settings, detailed results, and analysis. % Then, the results are reported and then report detailed experiment results. % This section describes the used datasets, results, and discussion for our model MUSE.
\subsection{Datasets}
\textbf{WMT14 En-Fr and En-De datasets} The WMT 2014 English-French translation dataset, consisting of $36M$  sentence pairs, is adopted as a big dataset to test our model. We use the standard split of development set and test set. We use \textit{newstest2014} as the test set and use \textit{newstest2012} +\textit{newstest2013} as the development set. Following \citet{gehring2017convolutional}, we also adopt a joint source and target BPE factorization with the vocabulary size of $40K$.
For medium dataset, we borrow the setup of \citet{vaswani2017attention} and adopt the WMT 2014 English-German translation dataset which consists of $4.5M$ sentence pairs, the BPE vocabulary size is set to $32K$. The test and validation  datasets we used are the same as \citet{vaswani2017attention}.

\textbf{IWSLT De-En and En-Vi datasets} Besides, we perform experiments on two small IWSLT datasets to test the small version of MUSE with other comparable models. The IWSLT 2014 German-English translation dataset consists of $160k$ sentence pairs. We also adopt a joint source and target BPE factorization with the vocabulary size of $32K$. The IWSLT 2015 English-Vietnamese translation dataset consists of $133K$ training sentence pairs. For the En-Vi task, we build a dictionary including all source and target tokens. The vocabulary size for English is $17.2K$, and the vocabulary size for the Vietnamese is $6.8K$. 

%  Following the fairseq repository~\citep{ott2019fairseq}, we use the BPE technique to preprocess the DE-EN, EN-DE, EN-FR datasets and remove BPE before the evaluation.   
% For the EN-VI task, .
% Following \cite{vaswani2017attention}, we report the tokenized BLEU score after compound splitting on the En-De dataset. 
% ZZY modify Experimental Settings暂时改完

\subsection{Experimental Settings}
%Since our model can help to build deeper models and cut down dimensions to maintain the model size accordingly Unlike the encoder, the decoder not only adds the output of the self-attention, Conv and FFN before the layer norm but also plus the output of the decoder context attention. 
% We build a model consisting of $12$ encoder layers and $12$ decoder layers. The hidden dimension is set to $384$ on DE-EN and EN-VI translation, and $768$ on EN-FR translation.
\subsubsection{Model}
For fair comparisons, we only compare models reported with the comparable model size and the same training data. We do not compare \citet{Wu_2019} because it is an  ensemble method. We build MUSE-base and MUSE-large with the parameter size  comparable to Transformer-base and Transformer-large. We adopt multi-head attention~\citep{vaswani2017attention} as implementation of self-attention in MUSE module. The number of attention head is set to 4 for MUSE-base and 16 for MUSE-large. We also add the network architecture built by MUSE-simple  in the similar way into the comparison.

MUSE consists of 12 residual blocks for encoder and 12 residual blocks for decoder, the dimension is set to 384 for MUSE-base and 768 for MUSE-large. The hidden dimension of non linear transformation is set to 768 for MUSE-base and 3072 for MUSE-large.

% The embedding layer is initialized by a normal distribution.   For the rest parameters, we use the default initialization method by \textit{pytorch}. Suppose  ${d_{in}}$ is the input dimension, the parameters  of the fully connected layer are initialized by an uniform distribution $(-1/\sqrt{d_{in}}, 1/\sqrt{d_{in}})$.

The MUSE-large is trained on 4 Titan RTX GPUs while the MUSE-base is trained on a single NVIDIA RTX 2080Ti GPU. The batch size  is calculated at the token level, which is called dynamic batching~\citep{vaswani2017attention}. 
We adopt dynamic convolution as the variant of depth-wise separable convolution. We tune the kernel size on the validation set. For convolution with a single kernel, we use the kernel size of $7$ for all layers. In case of dynamic selected kernels, the kernel size is $3$ for small kernels and $15$ for large kernels for all layers.

\subsubsection{Training}
The training hyper-parameters are tuned on the validation set. 

\textbf{MUSE-large}
For training MUSE-large, following \citet{ott2018scaling}, parameters are updated every $32$ steps. We train the model for $80K$ updates with a batch size of $5120$ for En-Fr, and train the model for ${30K}$ updates with a batch size of ${3584}$ for En-De. The dropout rate is set to $0.1$ for En-Fr and ${0.3}$ for En-De. We borrow the  setup of optimizer from \citet{wu2019pay} and use the cosine learning rate schedule with ${10000}$ warmup steps. The max learning rate is set to $0.001$ on  En-De translation and ${0.0007}$ on En-Fr translation. For checkpoint averaging, following \citet{wu2019pay}, we tune the average checkpoints for En-De translation tasks. For En-Fr translation, we do not average checkpoint but use the final single checkpoint.

\textbf{MUSE-base}
We train and test MUSE-base on two small datasets, IWSLT 2014 De-En translation and  IWSLT2015 En-Vi translation.  Following \citet{vaswani2017attention}, we use Adam optimizer with a learning rate of $0.001$. We use the warmup mechanism and invert the learning rate decay with warmup updates of $4K$. 
For the De-En dataset, we train the model for $20K$ steps with a batch size of $4K$. The parameters are updated every $4$ steps. The dropout rate is set to $0.4$. For the En-Vi dataset, we train the model for $10K$ steps with a batch size of $4K$. The parameters are  also updated every $4$ steps. The dropout rate is set to $0.3$.
We  save checkpoints every epoch and average the last $10$ checkpoints for inference.

\subsubsection{Evaluation}
During inference, we adopt beam search with a beam size of $5$ for De-En, En-Fr and  En-Vi translation tasks. The length penalty is set to 0.8 for En-Fr according to  the validation results, 1 for the two small datasets following the default setting of \cite{ott2019fairseq}. We do not tune beam width and length penalty but use the setting reported in \citet{vaswani2017attention}.
The BLEU\footnote{https://github.com/moses-smt/mosesdecoder/blob/master/scripts/generic/multi-bleu.perl} metric is adopted to evaluate the model performance during evaluation. 

\begin{table*}
\centering
\begin{tabular}{lcr}
\toprule
Model   & En-De & En-Fr \\
\midrule
% CNNSeq2seq~\citep{gehring2017convolutional}  & 1x & 1x & 25.2  & 40.5 \\
% Transformer~\citep{vaswani2017attention} &1x & 1x &28.4  & 41.0 \\
% RNMT+~\citep{Chen_2018} &1x & 2x  & 28.5 & 41.0 \\
% Evolved Transformer~\citep{so2019evolved} & 1x &1x &  29.8 & 41.3 \\
% Weighted Transformer~\citep{ahmed2017weighted} & 1x &1x & 28.9 & 41.4 \\
% Relative Transformer~\citep{Shaw_2018}  & 1x & 1x & 29.2 & 41.5 \\
% Transformer~\citep{ott2018scaling} & 1x & 1x & 29.3 & 43.2 \\
% DynamicConv~\citep{wu2019pay} &1x &1x & 29.7 & 43.2 \\
ConvSeq2seq~\citep{gehring2017convolutional}  & 25.2  & 40.5 \\
SliceNet~\citep{kaiser2017depthwise} & 26.1 &- \\
Transformer~\citep{vaswani2017attention}  &28.4  & 41.0 \\
Weighted Transformer~\citep{ahmed2017weighted}  & 28.9 & 41.4 \\
Layer-wise Coordination~\citep{NIPS2018_8019}  & 29.1 & - \\
Transformer (relative position)~\citep{Shaw_2018}   & 29.2 & 41.5 \\
Transformer~\citep{ott2018scaling} & 29.3 & 43.2 \\
Evolved Transformer~\citep{so2019evolved} &  29.8 & 41.3 \\
DynamicConv~\citep{wu2019pay}  & 29.7 & 43.2 \\

Local Joint Self-attention~\citep{fonollosa2019joint} &29.7  & 43.3 \\
\midrule
\textbf{MUSE-simple} & 29.8 & 43.2 \\
\textbf{MUSE}  & \textbf{29.9} & \textbf{43.5} \\
\bottomrule
\end{tabular}
\caption{MUSE-large outperforms all previous models under the standard training and evaluation setting on WMT14 En-De and WMT14 En-Fr datasets.}
% \caption{Comparisons between previous state-of-the-art models with comparable model size and training data size on the WMT14 En-De and  En-Fr translation task.}
\label{tab:enfr}
\vspace{-0.2cm}
\end{table*}

\begin{table}[t]
\centering
\begin{tabular}{lrrr}
\toprule
Model  & En-Vi & De-En  \\
\midrule
% NBMT~\citep{NBMT} &- & 28.1 & 30.1  \\
% SACT~\citep{SACT} &- & 29.1 &-  \\
% NP2MT \citep{feng2018neural} &-  &30.6 & 31.7 \\ % NEURAL PHRASE-TO-PHRASE MACHINE TRANSLATION
% Fixup \citep{zhang2019fixup} &1x & -  & 34.5\\
% DynamicConv \citep{wu2019pay} & 1x & - & 35.2 \\
% Macaron~\citep{lu2019understanding} &1x  & - &35.4 \\ 
% MAtt~\citep{zhang2019improving}  &2x &- &35.6 \\ 
NBMT~\citep{NBMT} & 28.1 & 30.1  \\
SACT~\citep{SACT} & 29.1 &-  \\
NP2MT~\citep{feng2018neural} &30.6 & 31.7 \\ % NEURAL PHRASE-TO-PHRASE MACHINE TRANSLATION
Fixup~\citep{zhang2019fixup}  & -  & 34.5\\
DynamicConv~\citep{wu2019pay}  & - & 35.2 \\
Macaron~\citep{lu2019understanding}  & - &35.4 \\ 
% ~\citet{zhang2019improving}  &2x &- &35.6 \\
Local Joint Self-attention~\citep{fonollosa2019joint}  &- &35.7 \\
\midrule
\textbf{MUSE-simple} & 30.7 & 35.8 \\
\textbf{MUSE} &\textbf{31.3} &\textbf{36.3}\\
\bottomrule
\end{tabular}
% \caption{BLEU scores of MUSE and state-of-the-art models on IWSLT  De-En, IWSLT  En-Vi translation datasets. }
\caption{MUSE-base outperforms previous state-of-the-art models on IWSLT  De-En translation datasets and outperforms previous models without BPE processing on IWSLT  En-Vi. }
\label{tab:mtsmall}
\end{table}

\subsection{Results}
%$$\subsection{Machine Translation}
%We compare various models (e.g. attention alone, convolution,depth-separable convolution, local attention, NAS, mix convolution)  on machine translation datasets.

As shown in Table~\ref{tab:enfr}, MUSE outperforms all previously  models on En-De and En-Fr translation, including both  state-of-the-art models of stand alone self-attention~\citep{vaswani2017attention,ott2018scaling}, and convolutional models~\citep{gehring2017convolutional,kaiser2017depthwise,wu2019pay}. This result shows that either self-attention or convolution alone is not enough for sequence to sequence learning. The proposed parallel multi-scale attention improves over them both on En-De and En-Fr.

% and establishes new state-of-the-art results  on the IWSLT De-En and En-Vi machine translation tasks. To be specific, MUSE achieves a 36.3 BLEU score on De-En translation and a 31.3 BLEU score on En-Vi translation. Furthermore, compared with the approaches with the similar model size, Fixup and Macaron, MUSE achieves almost 1.0 BLEU score improvement on De-En datast.   
% On the WMT En-Fr machine translation dataset, MUSE also achieves state-of-the-art results, with a BLEU score of 43.5, as shown in Table~\ref{tab:enfr}, this result shows that self attention  alone.
Compared to Evolved Transformer \citep{so2019evolved} which is constructed by NAS and also mixes convolutions of different kernel size, MUSE achieves 2.2 BLEU gains in En-Fr translation.

Relative position or local attention constraints bring improvements over origin self-attention model, but parallel multi-scale outperforms them.

MUSE can also scale to small model and small datasets, as depicted in Table~\ref{tab:mtsmall}, MUSE-base pushes the state-of-the-art from 35.7 to 36.3 on IWSLT De-En translation dataset.

It is shown in Table~\ref{tab:enfr} and Table~\ref{tab:mtsmall} that MUSE-simple which contains the  basic idea of parallel multi-scale attention achieves state-of-the-art performance on three major machine translation datasets.

\begin{table*}[t]
\centering
\begin{tabular}{lr}
\toprule
Model & BLEU \\
\midrule
MUSE & 36.3\\
MUSE-simple (without DepthConv) & 35.8 \\ \midrule
substitute DepthConv with Convolution (k=3) & 35.2 \\  
substitute DepthConv with Convolution (k=5) & 35.0 \\  
substitute DepthConv with Convolution (k=7) & 34.5 \\  \midrule
DepthConv without shared projection &34.9 \\
DepthConv single kernel (k=3) & 36.2 \\  
DepthConv single kernel (k=7) & 36.2 \\ 
DepthConv single kernel (k=15) & 36.0 \\  
DepthConv single kernel (k=31) & 35.8 \\
DepthConv single kernel (grow kernels among layers:3,7,15,31) & 35.9 \\
DepthConv dynamically selected kernel (k=3,15) & 36.3 \\
\bottomrule
\end{tabular}
\caption{Comparisons between  MUSE and its variants on the IWSLT 2015 De-En translation task.}
\label{tab:ablaton}
\end{table*}
\subsection{How do we propose effective parallel multi-scale attention?}
In this subsection we compare MUSE and its variants on  IWSLT 2015 De-En translation  to answer the question.
 
\textbf{Does concatenating self-attention with convolution certainly improve the model?}
To bridge the gap between point-wise transformation which learns token level representations and self-attention which learns representations of global context, we introduce convolution to enhance our multi-scale attention. As we can see from the  first experiment group of Table~\ref{tab:ablaton},   convolution is important in the parallel multi-scale attention. 
However, it is not easy to combine convolution and self-attention in one module to build better representations on sequence to sequence tasks. As  shown in the first line of both second and third group of Table~\ref{tab:ablaton}, simply learning local representations by using convolution or depth-wise separable convolution in parallel with self-attention  harms the performance. Furthermore, combining depth-wise separable convolution (in this work we choose its best variant dynamic convolution as implementation) is even worse than  combining convolution.

\textbf{Why do we choose DepthConv and what is the importance of sharing Projection of DepthConv and self-attention?}
We conjecture that convolution and self-attention both learn contextual sequence representations and they should share the point  transformation and perform the contextual transformation in the same hidden space. 
We first project  the input to a hidden representation  and perform a variant of  depth-wise convolution and self-attention transformations in parallel.  
The fist two experiments in third group of Table~\ref{tab:ablaton} show that  validating the utility of sharing Projection in parallel multi-scale attention, shared projection gain 1.4 BLEU scores over separate projection, and bring improvement of 0.5 BLEU scores over MUSE-simple (without DepthConv).

\textbf{How much is the kernel size?}
Comparative experiments show that the too large kernel harms performance  both for DepthConv and convolution. Since there exists self-attention and point-wise transformations, simply applying the growing kernel size schedule proposed in SliceNet~\citep{kaiser2017depthwise}  doesn't  work. Thus, we propose to use dynamically selected kernel size to let the learned network decide the kernel size for each layer.

\subsection{Further Analysis}

\paragraph{Parallel multi-scale attention brings time efficiency on GPUs}

The underlying parallel structure (compared to the sequential structure in each block of Transformer) allows MUSE to be efficiently computed on GPUs. For example, we can combine small matrices into large matrices, and while it does not reduce the number of actual operations, it can be better paralleled by GPUs to speed up computation.  Concretely, for each MUSE module,  we  first concentrate $W^Q,W^K,W^V$ of self-attention and $W_1$ of point feed-forward transformation into a single encoder matrix $W^{Enc}$, and then perform transformation such as self-attention, depth-separable convolution, and nonlinear transformation, in parallel, to learn multi-scale representations in the hidden layer. $W^O,W_2,W^{out}$ can also be combined a single decoder matrix  $W^{Dec}$. The decoder of sequence to sequence architecture can be implemented similarly.

In Table~\ref{tab:speed}, we conduct comparisons to show the speed gains with the aforementioned implementation, and the batch size is set to one sample per batch to simulate online inference environment. Under the settings, where the numbers of parameters are similar for MUSE and Transformer, about 31\% increase in inference speed can be obtained. The experiments use MUSE with $6$ MUSE-simple modules and Transformer with $6$ base blocks. The hidden size is set to $512$. %It is worth noticing that for the MUSE structure used in the main experiments, ideally a similar speedup can be witnessed if the computing device is powerful enough. However, such is not the case in our preliminary experiments. We also need to point out the implementation is far from fully optimized and the results are only meant to demonstrate the feasibility of the procedure.

% In Table~\ref{tab:speed}, we build MUSE by stacking $6$ MUSE-simple modules both for the encoder and the decoder, the dimension is set to $512$, results show that the parallel multi-scale setting has potential for speeding inference. 

\begin{table*}[ht]
\centering
\begin{tabular}{lr}
\toprule
Model  & Inference Speed (tokens/s) \\
\midrule
% Transformer-small &4.8M  &30 &1x & 1x \\ 
% MUSE-simple small  &4.8M  &30 &1.35x & 1.5x \\ \hline
% Transformer small &4.8M &1x & 1x \\ 
% MUSE-simple small  &4.8M  &30 &1.35x & 1.5x \\ \hline
% Transformer base &42.9M &35 &1x &  1 \\
% MUSE-simple base &42.8M &35 &1.15x & 1.35x \\
Transformer  &  132 \\
MUSE  & 173 \\ \midrule
Acceleration & 31\% \\
\bottomrule
\end{tabular}
\caption{The comparison between the inference speed  of  MUSE and Transformer.}
\label{tab:speed}
\end{table*}

\begin{figure}[ht]
\centering
\begin{minipage}[]{0.46\linewidth}  
\includegraphics[width=\textwidth]{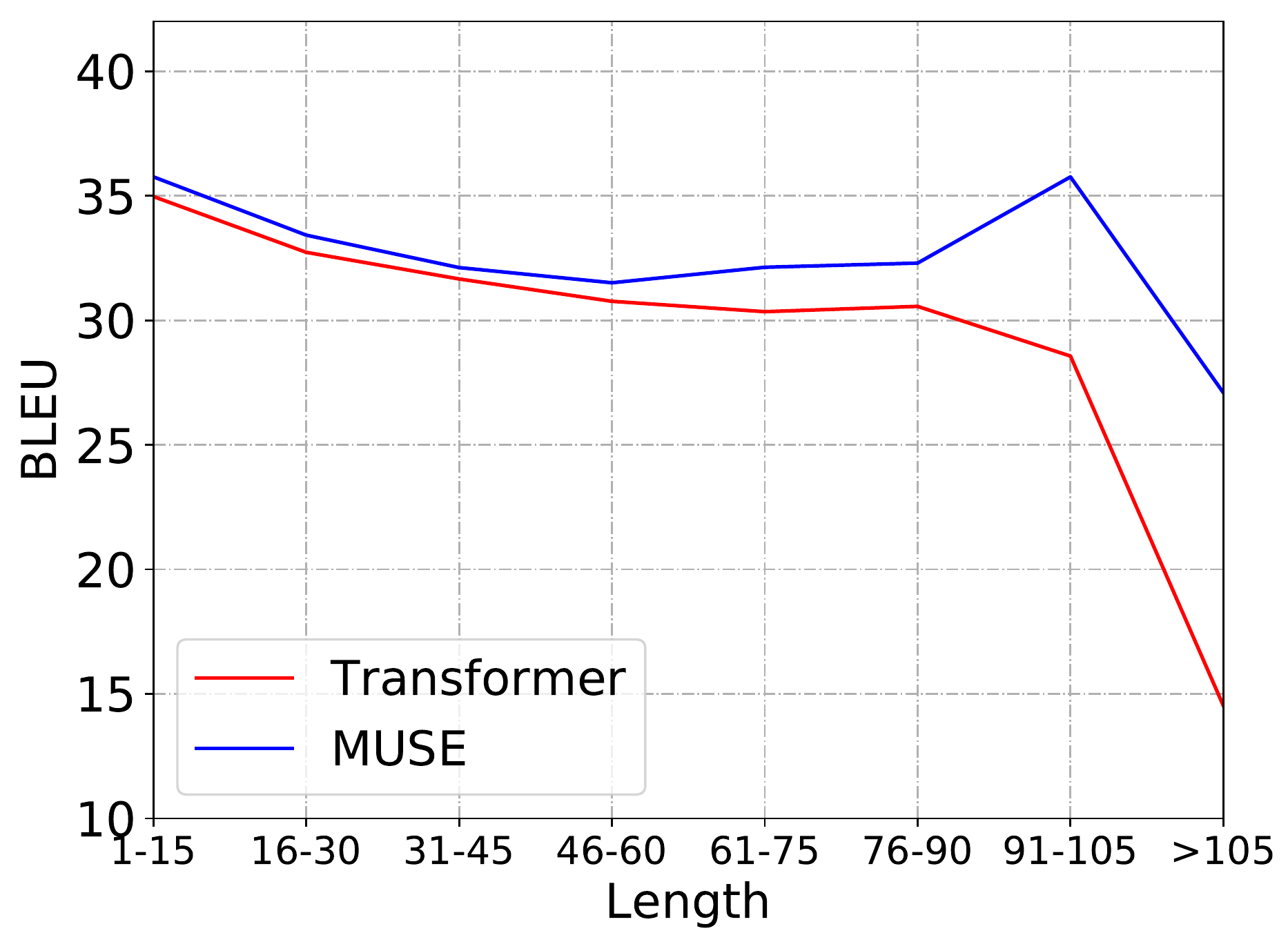}
\caption{BLEU scores of models on different groups with different source sentence lengths. The experiments are conducted on the De-En dataset. MUSE performs better than Transformer, especially on long sentences.}
\label{fig:bleu-length}
\end{minipage}
\qquad
\begin{minipage}[]{0.46\linewidth} 
\includegraphics[width=\textwidth]{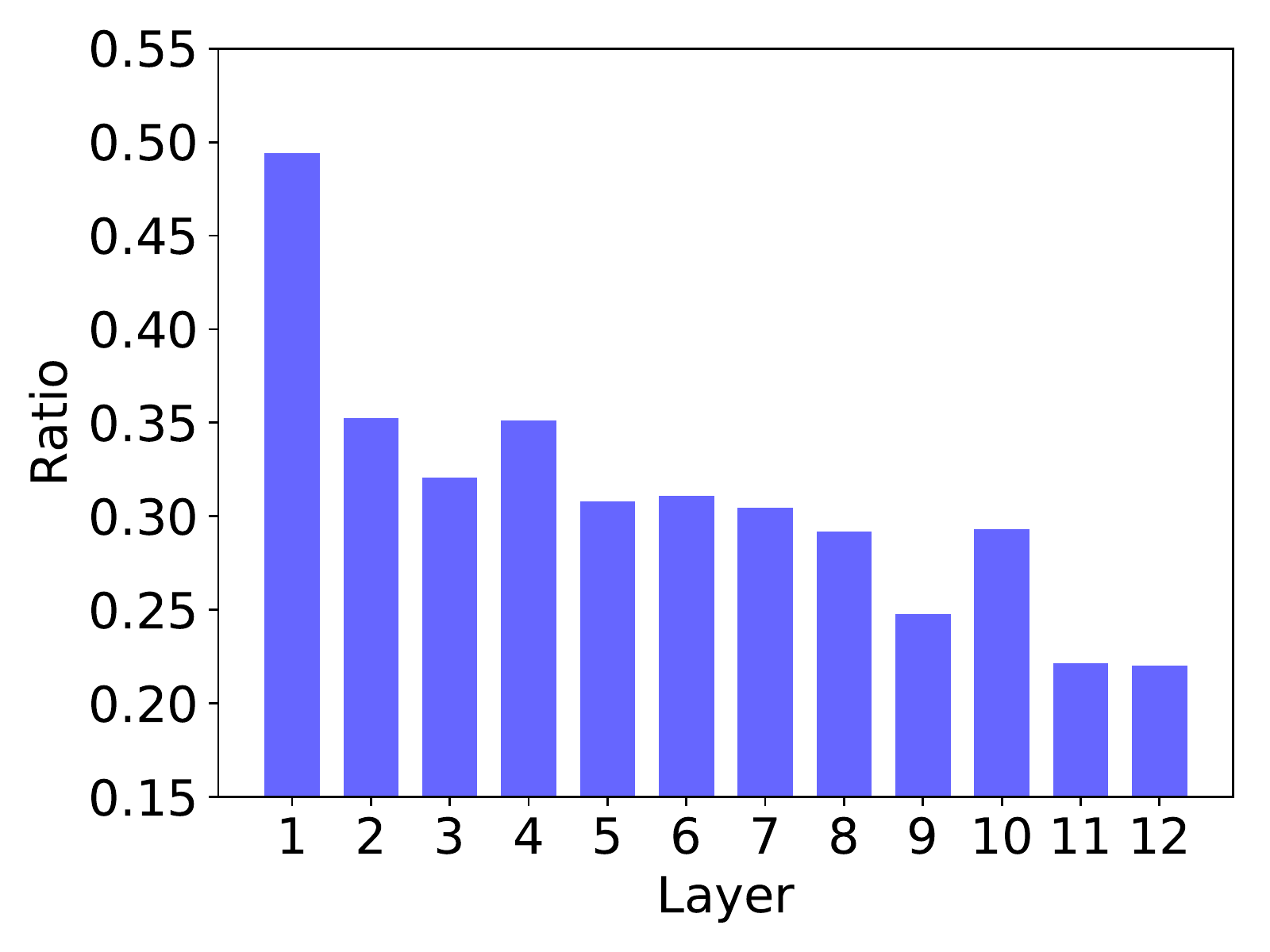}
\caption{Dynamically selected kernels at each layer: The blue bars represent the ratio between the percentage of the  convolution with  smaller kernel sizes and the percentage of the convolution with large kernel sizes.}
\label{fig:layer-k}
\end{minipage}
\end{figure}

\textbf{Parallel multi-scale attention generates  much better long sequence}
As demonstrated in Figure~\ref{fig:bleu-length}, MUSE generates better  sequences of various length than self-attention, but it is remarkably adept at generate long sequence, e.g. for sequence longer than 100, MUSE is two times better.

\textbf{Lower layers prefer local context and higher layers prefer more contextual representations}
MUSE contains multiple dynamic convolution cells, whose streams are fused by a gated mechanism. The weight for each dynamic cell is a scalar. Here we analyze the weight of different dynamic convolution cells in different layers.  
Figure~\ref{fig:layer-k} shows that as the layer depth increases, the weight of dynamic convolution cells with small kernel sizes gradually decreases. It demonstrates that lower layers prefer local features while higher layers prefer global features. It is corresponding to the finding in \citet{ramach2019standalone}.

%\begin{figure}[h]
%\centering
%\includegraphics[width=0.4\textwidth]{figures/deen-bleu-length.pdf}
%\caption{BLEU scores of on different groups with different source sentence lengths. The left and right figure is on the De-En and En-Vi dataset respectively.}
%\label{fig:bleu-length}
%\end{figure}

\textbf{MUSE not only gains BLEU scores, but also generates more reasonable sentences and increases the translation quality.}
We conduct the case study on the De-En dataset and the cases are shown in Table~\ref{tab:case_study} in Appendix.  In case 1, although the baseline transformer translates many correct words according to the source sentence, the translated sentence is not fluent at all.  It indicates that Transformer does not capture the relationship between some words and their neighbors, such as ``right'' and ``clap''. By contrast, MUSE captures them well by combining local convolution with global self-attention. In case 2, the cause adverbial clause is correctly translated by MUSE while transformer misses the word ``why'' and fails to translate it.

\section{Related Work}
% 先讲下sequence to sequence task
Sequence to sequence learning is an important task in machine learning. It evolves understanding and generating sequence. Machine translation is the touchstone of sequence to sequence learning. Traditional approaches usually adopt long-short term memory networks~\citep{sutskever2014sequence,ma-etal-2018-bag}  to learn the representation of sequences. However, these models either are built upon auto-regressive structures requiring longer encoding time or perform worse on real-world natural language processing tasks.
Recent studies  explore convolutional neural networks (CNN)~\citep{gehring2017convolutional} or self-attention ~\citep{vaswani2017attention} to  support high-parallel sequence modeling and does not require auto-regressive structure during encoding, thus bringing large efficiency improvements. They are strong at capturing local or global dependencies.
%  再讲下 长序列问题，局部attenition 没有在 seq2seq里解决这个问题, 不过不讲也没关系
% 现在 20:19 我先把 exp 写完再写conclusion

%讲下结合attention 和self attention 的相关研究
There are several studies on combining self-attention and convolution. However, they do not surpass both convectional and self-attention mechanisms.  \citet{sukhbaatar2019augmenting} propose to augment convolution with self attention by directly concentrating them in computer vision tasks.  However, as demonstrated in Table~\ref{tab:ablaton} there method does not work for sequence to sequence learning task. Since state-of-the-art models on question answering tasks still consist on self-attention and do no adopt ideas in QAnet~\citep{yu2018qanet}. Both self-attention~\citep{ott2018scaling} and convolution~\citep{wu2019pay} outperforms Evolved transformer by near 2 BLEU scores on En-Fr translation. It seems that learning global and local context through stacking self-attention and convolution  layers does not beat either self-attention or convolution models. In contrast, the proposed parallel multi-scale attention outperforms  previous convolution or self-attention  based models  on main translation tasks, showing its effectiveness for sequence to sequence learning.
%讲下local attention 解决长距离的论文

% In recent years, some researches are focusing on exploring simpler or stronger networks than Transformer. \citet{wu2019pay} propose a very lightweight convolution network, which is simpler and more efficient than Transformer. ~\citet{so2019evolved} adopt NAS, a neural architecture search method, to search for
% a better alternative to Transformer by fusing attention and convolutions together. In this work, we propose a multi-scale attention model by combing global features, local features, and token features together to improve the generalization ability.  

 % Furthermore, layer normalization and residual structure make it possible to build a very deep model, enabling the model with a powerful learning ability~\citep{Wang_2019}. Besides, ~\citet{zhang2019improving} proposes to build a deep transformer of 12 layers with a large number of parameters.
% some researchers also apply attention to computer vision tasks~\citep{Wang_2018,DBLP:journals/corr/abs-1904-09925}. Further, \citet{ramach2019standalone} claim that all convolutional nets can be replaced by self-attention in computer vision tasks to improve performance. 

\section{Conclusion and Future work}
%Transformer, a widely used sequence-to-sequence modeling network, has achieved promising results on natural language processing. 

% In this work, we rethink a fundamental question: ``Is attention really all you need?". 
Although the self-attention mechanism has been prevalent in sequence modeling, we find that attention suffers from dispersed weights especially for long sequences, resulting from the insufficient local information. 

To address this problem, we present Parallel Multi-scale Attention (MUSE) and MUSE-simple.  MUSE-simple introduces the idea of parallel multi-scale attention into sequence to sequence learning. And MUSE fuses self-attention, convolution, and point-wise transformation together to explicitly learn global, local and token level sequence representations. 
% Although empirical results prove the effectiveness of shared projection for combine convolution in MUSE, it is still unknown how this method affects model learning. 
Especially, we find from empirical results that the shared projection plays important part in its success, and is essential for our multi-scale learning.

Beyond the inspiring new state-of-the-art results on three major machine translation datasets, detailed  analysis and model variants also verify the effectiveness of MUSE. 

For future work, the parallel structure is highly extensible and provide many opportunities to improve these models. In addition, given the success of shared projection, we would like to explore its detailed effects on contextual representation learning. Finally, we are exited about future of parallel multi-scale attention and plan to apply this simple but effective idea to other tasks including image and speech.

\subsubsection*{Acknowledgments}
%We thank Xuancheng Ren for helpful discussions and comments. 
This work was supported in part by National Natural Science Foundation of China (No. 61673028).

\bibliography{iclr2020_conference}
\bibliographystyle{iclr2020_conference}

\newpage
\appendix
\section{Appendix}
\subsection{Case study}
\begin{table*}[ht]
\centering
%\footnotesize
\begin{tabular}{p{2cm}p{10cm}}
\toprule
\textbf{Case 1} & \\
\midrule
\textbf{Source} & wenn sie denken, dass die auf der linken seite jazz ist und die, auf der rechten seite swing ist, dann klatschen sie bitte. \\
\midrule
\textbf{Target} & if you think the one on the left is jazz and \textbf{\textcolor{blue}{the one on the right is swing, clap your hands.}} \\
\midrule
 \textbf{Transformer} & if you think it's jazz on the left, and \textbf{\textcolor{red}{those on the right side of the swing are clapping}}, please. \\
\midrule
\textbf{MUSE} & if you think the one on the left is jazz, \textbf{\textcolor{blue}{and the one on the right is swing, please clap.}} \\
\midrule\midrule
\textbf{Case 2} & \\
\midrule
\textbf{Source} & und deswegen haben wir uns entschlossen in berlin eine halle zu bauen, in der wir sozusagen die elektrischen verhältnisse der insel im maßstab eins zu drei ganz genau abbilden können. \\
\midrule
\textbf{Target} & and \textbf{\textcolor{blue}{that's why we decided}} to build a hall in berlin, where we could precisely reconstruct, so to speak, the electrical ratio of the island on a one to three scale. \\
\midrule
 \textbf{Transformer} & and so in berlin, \textbf{\textcolor{red}{we decided}} to build a hall where we could sort of map the electrical proportions of the island at scale one to three very precisely. \\
\midrule
\textbf{MUSE} & and \textbf{\textcolor{blue}{that's why we decided}} to build a hall in berlin, where we can sort of map the electric relationship of the island at the scale one to three very precisely. \\
\bottomrule
\end{tabular}
\caption{Case study on the De-En dataset. The red bolded words denote the \textbf{\textcolor{red}{wrong}} translation and blue bolded words denote the \textbf{\textcolor{blue}{correct}} translation. In case 1, transformer fails to capture the relationship between some words and their neighbors, such as ``right'' and ``clap''. In case 2, the cause adverbial clause is correctly translated by MUSE while transformer misses the word ``why'' and fails to translate it. }
\label{tab:case_study}
\end{table*}

\end{document}